\documentclass[11pt, onecolumn]{IEEEtran}

\usepackage[latin1]{inputenc}
\usepackage[T1]{fontenc}
\usepackage[brazil]{babel}

\newcommand{\figref}[1]{(Figure \ref{#1})}
\newcommand{\figrefs}[1]{Figure \ref{#1}}
\newcommand{\figrefp}[1]{Figures \ref{#1}}

\usepackage{graphicx,times,amsmath} 
\usepackage{graphics}

\IEEEoverridecommandlockouts

\begin{document}

\title{\ \\ \LARGE\bf Fuzzy Modeling of Electrical Impedance Tomography Image of the Lungs
\thanks{Harki Tanaka, Neli Regina Siqueira Ortega and Mauricio Stanzione Galizia are with Medical Informatics/LIM01, School of Medicine of University of São Paulo, Rua Teodoro Sampaio 115 Pinheiros CEP: 05405-000 São Paulo SP Brazil (phone: +55 11 30617682; fax: +55 11 30617382; emails: harkit@usp.br, neli@dim.fm.usp.br, maugalizia@yahoo.com.br).}
\thanks{João Batista Borges Sobrinho, and Marcelo Britto Passos Amato are with Department of Experimental Pneumology/LIM09, School of Medicine of University of São Paulo, Av. Dr. Arnaldo 455 Cerqueira César CEP:01246-903 São Paulo SP Brazil (phone:+55 11 30617361; fax:+55 11 3061 2492; emails: jbborges@unisys.com.br; amato@unisys.com.br).}}

\author{Harki Tanaka, Neli Regina Siqueira Ortega, Mauricio Stanzione Galizia,\\ João Batista Borges Sobrinho, and Marcelo Britto Passos Amato}

\maketitle

\begin{abstract}
Electrical Impedance Tomography (EIT) is a functional imaging method that is being developed for bedside use in critical care medicine. Aiming at improving the chest anatomical resolution of EIT images we developed a fuzzy model based on EIT's high temporal resolution and the functional information contained in the pulmonary perfusion and ventilation signals. EIT data from an experimental animal model were collected during normal ventilation and apnea while an injection of hypertonic saline was used as a reference . The fuzzy model was elaborated in three parts: a modeling  of the heart, a pulmonary map from ventilation images and, a pulmonary map from perfusion images. Image segmentation was performed using a threshold method and a ventilation/perfusion map was generated. EIT images treated by the fuzzy model were compared with the hypertonic saline injection method and CT-scan images, presenting good results in both qualitative (the image obtained by the model was very similar to that of the CT-scan) and quantitative (the ROC curve provided an area equal to 0.93) point of view. Undoubtedly, these results represent an important step in the EIT images area, since they open the possibility of developing EIT-based bedside clinical methods, which are not available nowadays. These achievements could serve as the base to develop EIT diagnosis system for some life-threatening diseases commonly found in critical care medicine. 
\end{abstract}

\section{Introduction}
In an Intensive Care Unit (ICU) the function of many organs are monitored using different devices: cardiac monitor, pulse oxymeter, arterial invasive pressure, among others. The lungs, despite the fact that they are vital organs, usually do not have a direct method of monitoring at the bedside. The physiological functions of the lungs, necessary for maintaining gas exchange, are ventilation (air distribution) and perfusion (blood circulation). EIT (Electrical Impedance Tomography) is a functional imaging method based on the distribution of electrical impedances within a volume, which is the human chest in this case, that has the potential to show these pulmonary functions. EIT devices are small, portable and do not cause harm to the patient \cite{Brown-2002}. Victorino et al. have been developing an EIT method to be used for bedside lung monitoring on the ICU. They have demonstrated that the variation in the EIT impedance images of the lungs are very well correlated with the regional changes in the air content within a region of interest \cite{Victorino-2004}. However, the cyclic movement of blood through the pulmonary vessels influences thoracic impedances, too. Consequently, the EIT images carry at least information about both, ventilation and perfusion. Eyüboglu has demonstrated that it is possible to separate the thoracic impedance variations due to blood perfusion and due to ventilation by using ECG-gated EIT images \cite{Eyuboglu-1989}. Nevertheless, even applying the ECG-gated method, the resultant images are difficult to interpret because of their insufficient anatomical resolution, and it is almost impossible to separate the heart from the lungs. In this paper we propose a fuzzy linguistic model to analyze EIT images in order to identify and separate the heart from the lung regions. In addition, we propose a method to map within the lung region its two main functions, that is, ventilation and perfusion.

\section{Materials and Methods}
\begin{figure}[htp]
\center
\includegraphics[width=3.1in]{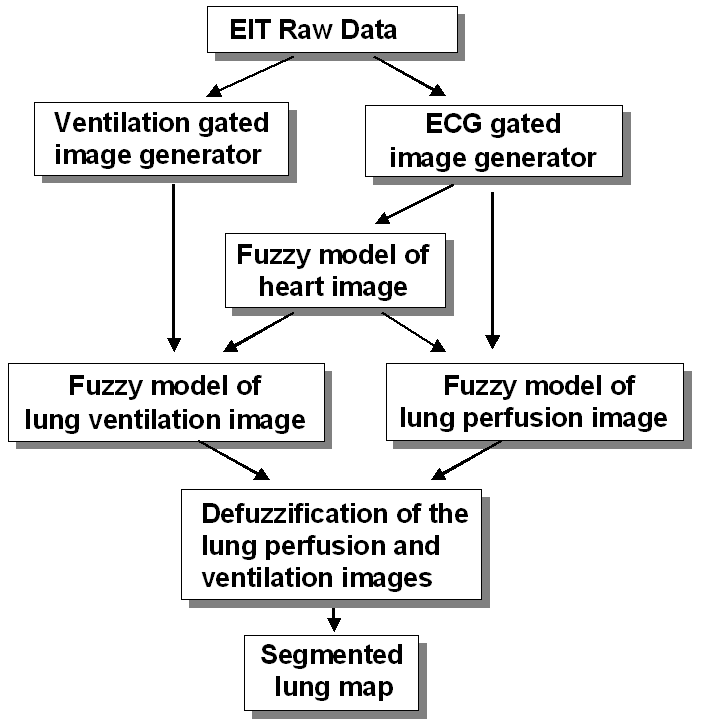}
\caption{EIT image analysis  structure.}
\label{esquemageral}
\end{figure}

Recently, fuzzy set theory has been used to deal with uncertainties present in health sciences and the results are very promising. It's aplicability covers a wide range of subjects, from epidemiological studies to diagnosing system development [4-7]\nocite{Pereira-2004}\nocite{Duarte-2006}\nocite{Massad-2002}\nocite{Massad-1999}. 
Our implementation of the EIT image treatment system employs the method of Mamdani and comprises software modules grouped in three steps:  EIT raw data acquisition and image generation step, fuzzy modeling step and image segmentation step \figref{esquemageral}. 

% Figure 1

\subsection{EIT images generation module}
\subsubsection{EIT Raw Data acquisition}The experiments were performed on an anesthetized healthy pig lying in the supine position, because the purpose of this work was to establish a fuzzy model under a well defined physiological condition. The animal was tracheostomized and maintained with continuous infusion of anesthetics. Controlled mechanical ventilation was delivered by a Servo 300 A (Siemens, Sweden). For the injection of hypertonic saline, a catheter was placed into the superior vena cava next to the entrance of the right atrium. ECG monitoring was performed using a Portal DX 2020 monitor (Dixtal, Brazil). EIT raw data were acquired using an impedance tomography device based on the ENLIGHT® technology (Dixtal, Brazil), capable of producing 50 images per second. Thirty-two electrodes were placed circumferentially (equally spaced) around the thorax just below the level of the axilla. An electrical current of 5 mA at 125 KHz was injected through a pair of electrodes while differential voltages were measured between the other non-injecting electrodes. Following this initial injedction, the electrical current was then injected sequentially via the next pair of electrodes and repeated until all electrodes had served for current injection. The data of one complete turn produced a so called frame, and were saved in a raw data file for a later processing. Ventilation was delivered by a mechanical ventilator in the pressure control mode. At this mode, the operator sets at lest the following three different parameters: the minimum pressure (Positive End-Expiratory Pressure = PEEP), the maximum pressure (plateau pressure) and the respiratory rate. The PEEP was set at 18 cmH2O and the plateau pressure at 28 cmH2O. Respiratory rate was constant at 20 cycles per minute. Seven separate EIT data sets were acquired together with ECG waveforms. Each EIT raw data set consisted of a first set of 10,000 frames during normal ventilation. Thereafter the ventilation of the animal was put on hold for a pre-defined period of time (apnea), while the pressure within the airways was maintained at the PEEP level. During this state of apnea, 5,000 images were acquired. Then 5 ml of a hypertonic (as compared to lung tissue of blood) solution of NaCl 20\% were injected quickly through the catheter and another 5,000 more images were acquired. Due to its high conductivity, the hypertonic saline solution acts as an EIT contrast agent. The resulting contrast injection images were used as a reference for the fuzzy model evaluation (as will be explained below).

\subsubsection{Generation of ECG-gated images and ventilation-gated images}For each set of 20,000 frames of EIT raw data, the synchronously recorded ECG waveforms were used to reconstruct a sequence of time-varying ECG-gated perfusion images. From the ECG waveform, a trigger pulse was produced at the rising edge of each R-wave, which is the moment when the ventricles are electrically stimulated just prior to their contraction, thus this point marks the beginning of the systolic part of the cardiac cycle. Following the detection of the R-wave, a block of frames was stored until the occurrence of the subsequent R-wave (1 cardiac cycle). This process was repeated until 100 complete cardiac cycles were stored and one ``mean cardiac cycle'' could be generated from them. Further ``mean cardiac cycles'' were generated from subsequent data sets of one hundred cardiac cycles. This sequence of mean cycles was processed using an image reconstruction algorithm, thus obtaining a series of ECG-gated images, in which the effects of ventilation is attenuated. A similar process was performed in order to generate ventilation-gated images, using at this time the beginning of each respiratory cycle as the trigger signal.  A block of frames between two subsequent trigger signals (1 respiratory cycle) was stored and this process was repeated sequentially, until 12 separate respiratory cycles were stored. Then one ``mean respiratory cycle'' from each block of 12 cycles was generated. This process was repeated for all subsequent data sets. The resultant sequence of ``mean respiratory cycles'' was processed using an image reconstruction algorithm, thus obtaining a series of ventilation-gated images, in which the effects of lung perfusion is attenuated. For image reconstruction we used the black-box back-projection algorithm developed by Lima and collaborators \cite{Aya-2005}, \cite{Lima-2005}.

\subsection{Fuzzy model to EIT images generation}
Each EIT image is formed by a matrix containing 32x32 pixels. The fuzzy modeled image was obtained by running the model once for each pixel, requiring 1024 runs to form one modeled image. All fuzzy linguistic models developed for this study applied the Mamdani inference procedure and the center of area defuzzification method, and were based on expert experience in EIT chest image analysis.
 
\subsubsection{Heart fuzzy model}The fuzzy linguistic model for the heart has three antecedent variables in its propositions: normalized perfusion amplitude, normalized time delay (TD) and pixel position, all of them were derived from ECG gated images; and one consequent variable: the possibility that the pixel carries the heart information (heart possibility). The pixel position is derived from the pixel order. The pixel of order 1 is located at the upper-left corner and the pixel of order 1024 is located at the lower-right corner of the image. Pixel orders from 1 to 512 belong to the anterior  region of the image, while pixel orders from 513 to 1024 belong to the posterior  region of the image.

\subsubsection{Lung perfusion fuzzy model}The lung perfusion linguistic fuzzy model has two antecedent variables: normalized perfusion amplitude and normalized heart possibility; and one consequent variable: the possibility that the pixel carries lung perfusion information. The defuzzified output obtained from running the heart possibility model was normalized in the interval [0,1] to obtain the normalized heart possibility. This fuzzy model is composed by nine inference rules. 

\subsubsection{Lung ventilation fuzzy model}As in the perfusion lung model, the lung ventilation fuzzy model has two antecedent variables: normalized ventilation amplitude and normalized heart possibility; and one consequent variable: the possibility that the pixel carries lung ventilation information. The normalized ventilation amplitude was derived from ventilation-gated images. It was selected from the ventilation-gated images an image data segment comprising at least one respiratory cycle. Then, for each pixel, the amplitudes were calculated as the difference of maximum and minimum impedance values in the selected respiratory cycles. The pixel amplitudes were normalized in relation to the highest amplitude. This fuzzy model is composed also by nine inference rules.

\subsection{Median fuzzy modeled lung images} 
    The fuzzy models, as previously described and depicted in \figrefs{esquemageral}, were run for each of the seven EIT raw data sets acquired in the present experiment, totalizing seven lung perfusion images and seven lung ventilation images. For evaluation purposes, it was generated two representative images: median lung perfusion image and median lung ventilation image, both resultants from the pixel-by-pixel median of the seven images, respectively. 

\subsection{Hypertonic saline images}
\begin{figure}[htp]
\centerline{\includegraphics[width=3.1in]{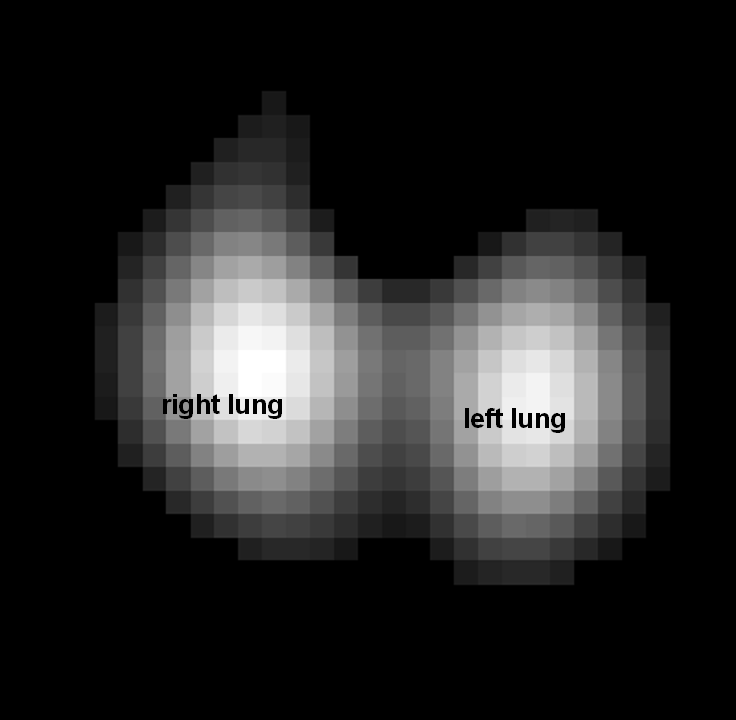}}
\caption{Lung perfusion reference image, obtained by hypertonic saline injection.}
\label{injectionlungreference}
\end{figure}

    The hypertonic saline solution acts as an EIT image contrast, because it is much more conductive than blood, generating a much higher signal than physiological perfusion. During the data acquisition protocol, a set of images were made during apnea (without ventilation) in which a hypertonic saline solution (NaCl 20\%) was injected through a catheter inserted inside the right atrium of the heart. A series of images were made just after the injection. These images show the pattern of the saline flow inside the chest: from the right atrium through the right ventricle, pulmonary arteries and lung vessels, until the recollection through the pulmonary veins to the left atrium. Experts in EIT images, observationally, selected a representative lung saline injection image and it was used as the reference image of the lung perfusion image \figref{injectionlungreference} \cite{Frerichs-2002}.

%Figure 2

\subsection{Segmented Lung perfusion and lung ventilation images}
    For evaluation purposes and in order to partition the modeled images in regions of practical interests, a segmented image was generated. The method used for segmentation was the threshold of the modeled images.  The images were submitted to threshold values, generating two images, one representing the lung perfusion map and the other representing the lung ventilation map. This methodology consists in a defuzzification procedure of the two fuzzy lung images, in a theoretical point of view. A total lung map was generated as the classical union of the two previous ones.
\subsection{Fuzzy modeled images evaluation}
    The lung perfusion image was compared with the lung saline injection image, considered as the reference image \figref{injectionlungreference}.  Two variables were calculated: a) sensibility, defined as the  number of pixels that belonged at the same time to the lung perfusion map and the reference image, divided by the number of pixels in the reference image; b) specificity, defined as the number of pixels that, at the same time, did not belong either to the perfusion map or to the reference image, divided by  the total  number of pixels that did not belong to the reference image. Both sensibility and specificity, when closer to the value one (1), indicate a better match between reference and target image.
    To generate the segmented reference image, a fixed threshold of 0.1 was used. Then, sensibilities and specificities were calculated for each perfusion map, with varying thresholds from 0.1 to 1, stepped by 0.05, in order to build the ROC curve.
    
\subsection{Qualitative comparison of segmented image and CT-scan image}
    The aim of image segmentation is to partition the image into several constituent components. In our study, the major components of interest are lung perfusion map, lung ventilation map and heart region. Despite the difference of purposes of the two imaging methods, we chose the Computerized Tomography scan (CT-scan) image as the anatomic gold standard for a qualitative comparison with the fuzzy modeled segmented image.
    Using a heuristically determined threshold of 0.31 for the ventilation image and 0.28 for the perfusion image, we have obtained a segmented image with three regions of interest: ventilation to perfusion matched region, predominantly ventilated region and predominantly perfused region. This segmented image was compared with a CT-scan image of a pig, acquired in a different animal but at similar conditions of ventilation and axial level.

\newpage
\section{Results}
\begin{figure}[tp]
\centerline{\includegraphics[width=3.1in]{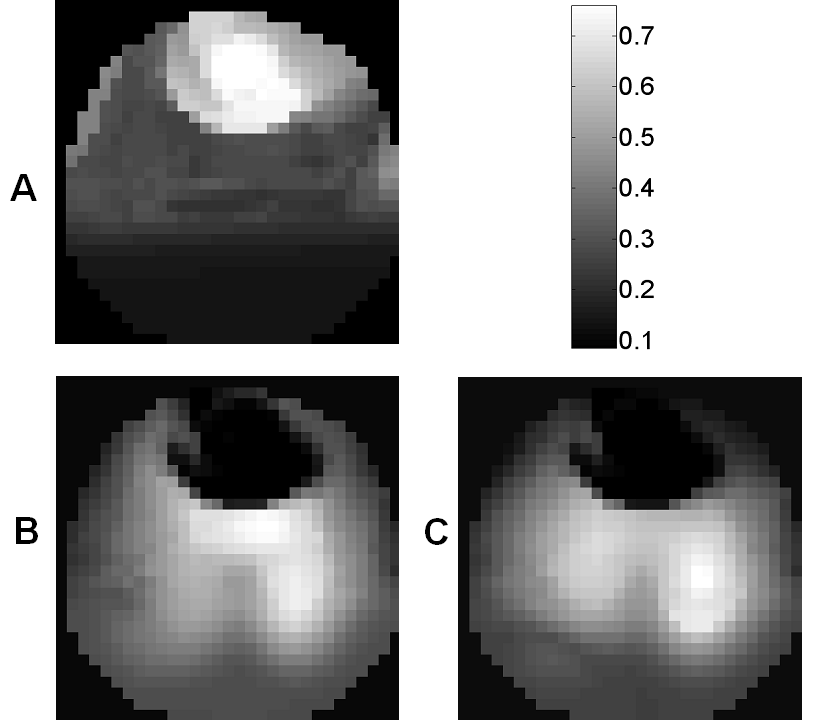}}
\caption{Gray scale map of fuzzy modeling results: A) median heart possibility map.  B) median lung perfusion possibility map. C) median lung ventilation possibility map.}
\label{heartperfusionventilation}

\end{figure}

In all the seven cases the fuzzy modeled heart images are identified in the anterior region  (\figrefs{heartperfusionventilation}-A) as expected by the a priori anatomical knowledge when the animal is in supine position. The lung perfusion possibility map (\figrefs{heartperfusionventilation}-B) shows a clear subtraction of the heart image from the original perfusion image. The same heart subtraction occurs with ventilation possibility image (\figrefs{heartperfusionventilation}-C).

\begin{figure}[tbp]
\centerline{\includegraphics[width=4in]{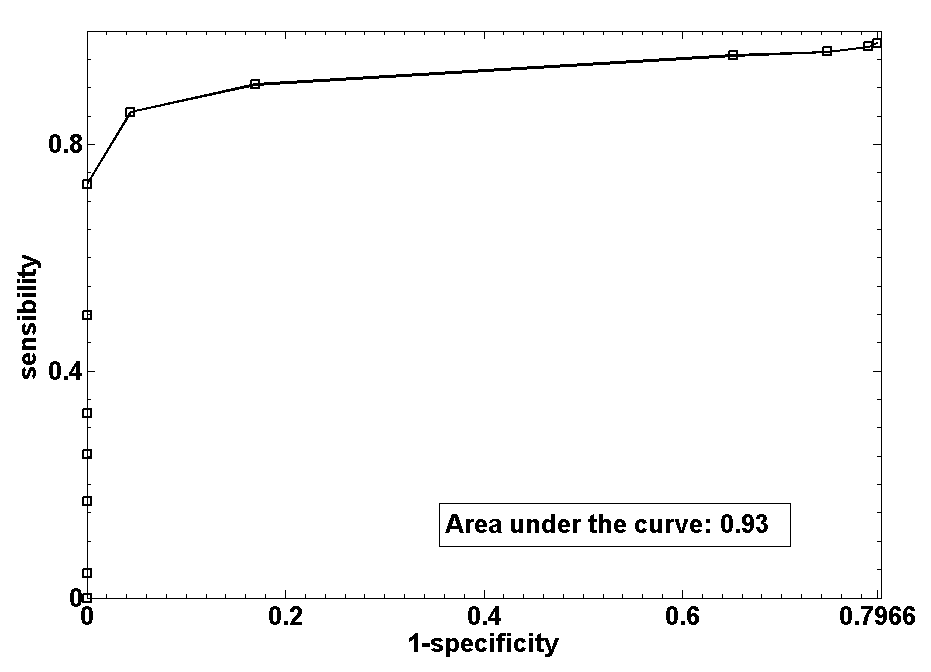}}
\caption{ROC curve for lung perfusion map evaluation.}
\label{ROCwitharea}
\end{figure}

A ROC curve (sensibility vs. 1-specificity) was plotted comparing the lung perfusion image with saline injection reference image \figref{ROCwitharea}. The area under the ROC curve is 0.93.  

\begin{figure}[tbp]
\centerline{\includegraphics[width=3.5in]{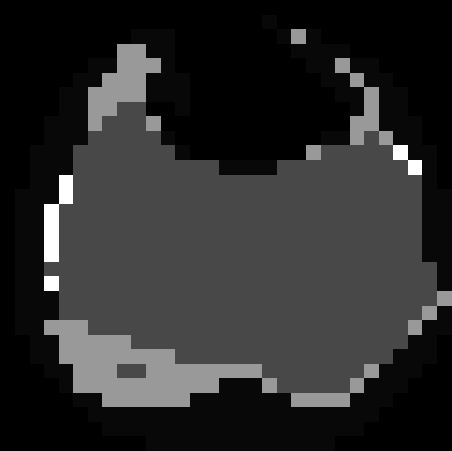}}
\caption{Segmented lung map composed by three regions: dark gray: perfusion and ventilation match; light gray:  predominantly perfused region; and white:  predominantly ventilated region.}
\label{segmentedlung}
\end{figure}

On the segmented lung map \figref{segmentedlung} the predominantly perfused regions (light gray) are founded at the anterior region and mainly at the lower left region of the lung. The matched area (dark gray) is concentrated at the middle region and a thin predominantly ventilated slice (white) appears at the lung periphery.  The heart region is clearly subtracted from the lung map. 

\begin{figure}[tbp]
\centerline{\includegraphics[width=3.5in]{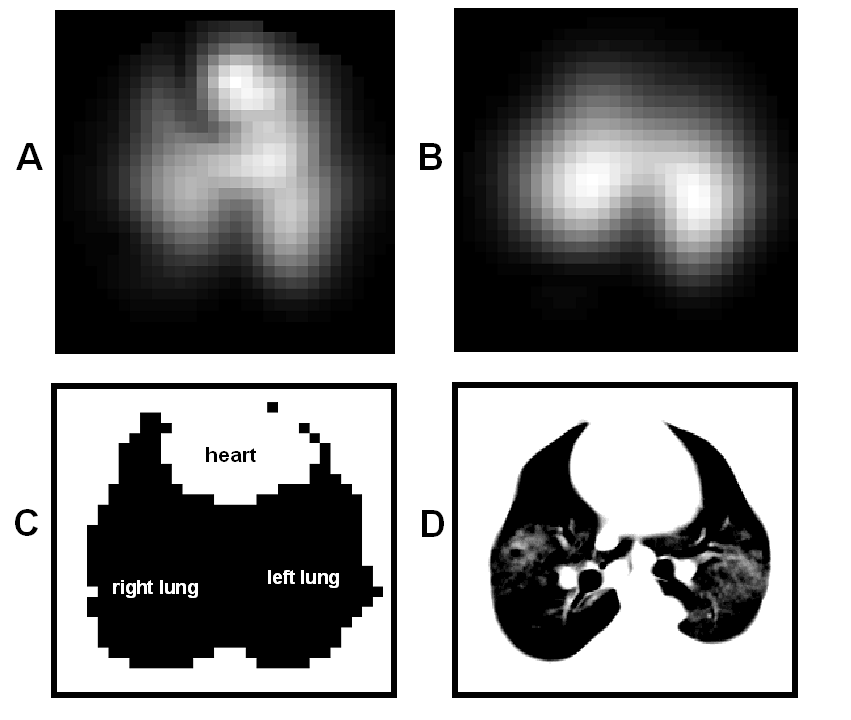}}
\caption{The EIT image of the lung before fuzzy analysis: lung perfusion  (A), lung ventilation (B), the total lung map through the model (C) and CT-scan image (D), for qualitative comparison.}
\label{comparison}
\end{figure}

The comparison between the total lung map found through the model (\figrefs{comparison}-C) and the CT-scan image (\figrefs{comparison}-D) shows a qualitative anatomic similarity of both the heart and lung structures. In the CT-scan image there is a clear separation between the left lung and the right lung, but in the total lung map the lungs are presented as one single region, without left and right delineation. \figrefp{comparison}-A and \ref{comparison}-B show the original EIT images of perfusion and ventilation respectively, before the fuzzy analysis. 

%Figure 3
%Figure 4

%Figure 5

%Figure 6

\newpage
\newpage
\section{Discussion}EIT is a medical imaging technology that has evolved from tracking pulmonary ventilation \cite{Victorino-2004} to the study of lung perfusion \cite{Eyuboglu-1989}, \cite{Frerichs-2002}. However, combining the functional information contained in both ventilation and perfusion images to process clinically useful additional information remains a challenge. One important step on the way to achieve this objective is to construct a pulmonary mask which delineats the lung's anatomic boundaries. Other lung imaging methot like scintigraphy relies on the application of radioactive tracers. In contrast, EIT is a non-invasive and real time methodology with no known side effects. Our fuzzy modeling of EIT images uses simultaneously acquired information from both, lung ventilation and perfusion. The results have shown that this method can delineate lung regions of particular clinical importance that have previously not been visible.

The heart fuzzy model proposed in this study clearly identified the heart region. When compared to a CT-scan image of the animal thorax, the modeled heart showed a good correspondence in position and in shape (\figrefp{heartperfusionventilation}-A and \ref{comparison}-C). 

In this work we first identified the heart region and then, by heart subtraction, we obtained the perfused and ventilated lung regions (\figrefp{heartperfusionventilation}-B and \ref{heartperfusionventilation}-C). The anatomical a priori knowledge that the heart is normally situated in the anterior portion of the chest made it possible to use the pixel position as one of the antecedent variables of the fuzzy model. The use of pixel position resulted in a lower heart possibility at the posterior region of the image, as can be observed in \figrefs{heartperfusionventilation}-A. In contrast, the simple parameter of pixel position cannot be used to identify the lung regions since pulmonary ventilation and perfusion occurs in the anterior and posterior regions of the thorax at the same time.

The ROC curve \figref{ROCwitharea} was plotted to evaluate the quality of the lung perfusion map in comparison to the reference method \figref{injectionlungreference}. The area under the curve was 0.93, showing a good agreement between them. Accordingly to this ROC curve the optimal threshold range for image segmentation lies between 0.3 and 0.4.  If the threshold is above 0.4 the resulting segmented image has a specificity of 1, which means that the whole image represents the perfused lung.   

Comparing the segmented total EIT lung map with the CT-scan image \figref{comparison}, a qualitative anatomic similarity of these lung structures can be noted, with the exception that on the CT-scan the right and left lung are separated by a central region not seen on the EIT-derived images. On the EIT segmented image both lungs are fused together in the middle of the thorax.  We can point out one factor that may be responsible for this difference: the low spatial resolution of the EIT method.

Observing the lung-segmented map \figref{segmentedlung} we noted that there is a main area (dark gray pixels) where the ventilation and perfusion were well matched. However, there are two main regions with perfusion dominance (light gray pixels): one at the upper-left quadrant and the other at the lower-left quadrant of the image. The two possible reasons for this mismatch of ventilation and to perfusion are: 1) in the anterior chest region the alveoli are overdistended and therefore receive less ventilation, and 2) in the posterior region the lung tend to collapse (atelectasis), thus they receive no ventilation . Besides that, in the lung periphery we found two thin predominantly ventilated regions (white pixels), which may be explained by the fact that the organ receives less perfusion at its periphery. These effects are particularly true at the high PEEP situation, as the 18  cmH2O PEEP used in the experiment of this work \cite{Hakim-1987}.

Although the results presented refer to the data obtained from one single animal experiment, the model was applied to other animals EIT images, providing consistent results. However, the quantitative analysis was not possible due to the lack of lung reference images of these animals (hypertonic saline injection reference image). 

The EIT image reconstruction algorithm assumes that the measurement electrodes are placed equidistantly and at the same transversal plane of the thorax.  In practice this may not always be true as long as individual electrodes are placed manually. Thus, there might be a considerable positioning error. Due to the lack of another applicable gold standard, in this study we used EIT hypertonic saline injection images as an anatomical reference. These reference images were produced by the same algorithm that generated the modeled fuzzy images  and therefore be subjected to the same positioning errors. In addition, since the aim of this study was to develop an EIT imaging tool that is based on fuzzy models, the analyzed data were acquired only on healthy pigs. Thus, the robustness of the method should also be tested in abnormal lung conditions. Nevertheless, the innovative aspects of our work are the use of temporal information for the heart structure delineation and the use of two distinct functional maps of lung ventilation and perfusion to delineate regions of particular clinical interest.

Finally, we want to point out that the present model provides the foundation to start dealing with clinically important problems such as disturbances of the matching of ventilation and perfusion, thromboembolism, atelectasis and cardiac output monitoring. In addition, the EIT method presented is non-invasive and low cost when compared to other image methods available today (i.e. CT-scan and PET).

\section{Conclusions}
The method for EIT image fuzzy modeling presented in this study provided very good results when compared with the reference methods. Besides an anatomic image similar to CT-scan, separating heart and lung also provided a segmented image in which the mapping of the ventilation and perfusion pulmonary functions were observed. The model provided new lung structure delineation based on pulmonary functions not available before in the original EIT images.
These achievements could serve as the base for development of an EIT based clinical tool for the diagnosis of some critical diseases commonly prevalent in the critical care units.

\section*{Acknowledgment}
This work was supported by the CNPq under grant no. 309135/2003-6 and by FAPESP under grant no. 01/053034.

\bibliographystyle{IEEEtran}
\bibliography{IEEEabrv,IEEEexample}

\end{document}